\documentstyle[aclap,psfig]{article}

\title{\vspace{-0.5in}Resolving Anaphors in Embedded Sentences}
\author{Saliha Azzam\\
University of Sheffield\\
Department of Computer Science\\
Regent Court\\
211 Portobello Street\\
Sheffield S1 4DP U.K.\\
{\tt S.Azzam@dcs.shef.ac.uk}}

\begin{document}
\maketitle
\vspace{-0.5in}
\begin{abstract}
We propose an algorithm to resolve anaphors, tackling mainly the problem
of intrasentential antecedents. We base our methodology on the fact that such 
antecedents are likely to occur in embedded sentences. Sidner's focusing mechanism 
is used as the basic algorithm in a more complete approach. The proposed algorithm
has been tested and implemented as a part of a conceptual analyser, mainly to
process pronouns. Details of an evaluation are given.
\end{abstract}

\section{Introduction}
Intrasentential antecedents, i.e., antecedents occurring in the same sentence
as the anaphor, are a crucial issue for the anaphora resolution method. The
main problem is to determine the constraints that intrasentential
phrases must respect in anaphoric relations. These constraints are used to
determine relations between a given anaphor and its antecedents. Until now,
this kind of constraint has been tackled mainly in terms of syntactic aspects,
see 
\cite{Lappin+Leass:GT} \cite{Merlo:ACL93} and \cite{Hobbs:ACL85}. We propose to
consider new kinds of criteria that combine semantic restrictions with sentence structure. 

One of these criteria is, for example, the way in which the verb meaning influences the sentence structure, 
then the way in which the sentence structure influences the anaphoric relations
between intrasentential phrases. The structure we studied is the embedded sentence
structure. Indeed, an important assumption we have made is that embedded
sentences favour the occurrence of intrasentential antecedents. We exploit the focusing mechanism proposed by Sidner \cite{Sidner:ACL79} \cite{Sidner:ACL81} 
\cite{Sidner:ACL83} extending and refining her algorithms. The algorithm is 
designed for anaphors generally,
even if we focus mainly on pronouns in this paper. Indeed, the distinction
between different kinds of anaphors is made at the level of anaphor interpretation rules. These resolution rule aspects will not be developed here, though.
They have been developed in the literature, e.g., see \cite{Carter:ACL87},
and \cite{Sidner:ACL81} \cite{Sidner:ACL83}. We focus more on the mechanisms that handle these different kinds of rules.

We first present how intrasentential antecedents occur in embedded sentences.
We recall the main ideas of the focusing approach, then expand on the main
hypotheses which led to the design of the anaphora resolution algorithm.

\section{Intrasentential Antecedents}
\subsection{Embedded sentences and elementary events}
An embedded sentence contains either more than one verb or a verb and
derivations of other verbs (see sentence 1 with verbs {\it said} and 
{\it forming}).
\begin{quote}
1) Three of the world's leading advertising groups, Agence Havas S.A. of
France, Young \& Rubicam of the U.S. and Dentsu Inc. of Japan, said they are
forming a global advertising joint venture.
\end{quote}
Broadly speaking embedded sentences concern more than one fact. In sentence 1
there is the fact of saying something and that of forming a joint venture. We
call such a fact an elementary event (EE hereafter). Thus an embedded sentence
will contain several EEs.

Factors that influence embedded sentences are mainly semantic features of verbs. For
example the verb {\it to say}, that takes a sentence complement favours an introduction
of a new fact, i.e., ``to say something'', and the related fact. There are other
classes of verbs such as {\it want to}, {\it hope that}, and so on.  
In the following, subordinate phrases, like relative or causal sentences, will be 
also considered as embedded ones.

\subsection{Embedded sentences with intrasentential antecedents}

First of all, we will distinguish the Possessive, Reciprocal and Reflexive
pronouns (PRR hereafter) from the other pronouns (non-PRR hereafter).

On the basis of 120 articles, of 4 sentences on average, containing 332
pronouns altogether, we made the following assumption (1):
\begin{description}
\item[Assumption :]
non-PRR pronouns can have intrasentential antecedents, only if these pronouns
occur in an embedded sentence.
\end{description}
The statistics below show that of 262 non-PRR pronouns, there are 244 having
intrasentential antecedents, all of which occur in embedded
sentences and none in a ``simple'' sentence.  The remaining 18 non-PRR
pronouns have intersentential antecedents.\\
\begin{tabbing}
123456789012444444444444444444444 \= \kill
Pronouns \>                      332\\
non-PRR  \>                      262\\
With intrasentential antecedents  \>   244\\
in an embedded sentence \> \\
With intrasentential in a simple  \>      0\\
sentence \> \\
With intersentential antecedents \>       18\\
\end{tabbing}

Our assumption means that, while the PRR pronouns may find their antecedents in
an non embedded sentence (e.g., sentences 2 and 3) the non-PRR pronouns can not.
\begin{quote}
2) Vulcan made  {\it its} initial Investment in Telescan in May, 1992.\\
3) The agencies HCM and DYR are {\it themselves} joint ventures.\\
\end{quote}
Without jumping to conclusions, we cannot avoid making a parallel with the
topological relations defined in the binding theory (Chomsky, 1980), between
two coreferring phrases in the syntactic tree level. Assumption 1 redefines
these relations in an informal and less rigorous way, at the semantic
level, i.e., considering semantic parameters such as the type of verbs that
introduce embedded sentences.

\subsection{Using Sidner's Focusing Approach}

To resolve anaphors one of the most suitable existing approaches when dealing
with anaphor issues in a conceptual analysis process is the focusing approach
proposed by Sidner. However, this mechanism is not suitable for
intrasentential cases. We propose to exploit its main advantages in order to
build our anaphora resolution mechanism extending it to deal also with
intrasentential antecedents.

We describe the main elements of the focusing approach that are necessary to 
understand our method, without going into great detail, see \cite{Sidner:ACL79} \cite{Sidner:ACL81} \cite{Sidner:ACL83}. Sidner proposed a methodology, modelling 
the way "focus" of attention and anaphor resolution influence one another. Using 
pronouns reflects what the speaker has focused on in the previous sentence, so that 
the focus is that phrase which the pronouns refer to. The resolution is organised 
through the following processes:
\begin{itemize}
\item The expected focus algorithm that selects an initial focus called the
  "expected focus". This selection may be ``confirmed'' or ``rejected'' in
  subsequent sentences. The expected focus is generally chosen on the basis of
  the verb semantic categories. There is a preference in terms of thematic
  position: the ``theme'' (as used by Gruber and Anderson, 1976 for the
  notion of the object case of a verb) is the first, followed by the goal, the
  instrument and the location ordered according to their occurrence in the
  sentence; the final item is the agent that is selected when no other role
  suits.
\item The anaphora interpreter uses the state of the focus and a set of
  algorithms associated with each anaphor type to determine which element of the
  data structures is the antecedent. Each algorithm is a filter containing
  several interpretation rules (IR).
  
  Each IR in the algorithm appropriate to an anaphor suggests one or several
  antecedents depending on the focus and on the anaphor type.
\item An evaluation of the proposed antecedents is performed using different
  kinds of criteria (syntactic, semantic, inferential, etc.)
\item The focusing algorithm makes use of data structures, i.e., the focus
  registers that represent the state of the focus: the current focus (CF)
  representation, alternate focus list (AFL) that contains the other phrases
  of the sentence and the focus stack (FS). A parallel structure to the CF is
  also set to deal with the agentive pronouns. The focusing algorithm updates
  the state of the focus after each sentence anaphor (except the first
  sentence). After the first sentence, it confirms or rejects the predicted
  focus taking into account the results of anaphor interpretation. In the case
  of rejection, it determines which phrase is to move into focus.
\end{itemize}

This is a brief example (Sidner 1983) :
\begin{enumerate}
\item[a] Alfred and Zohar liked to play baseball.
\item[b] {\it They} played it every day after school before dinner.
\item[c] After {\it their} game, Alfred and Zohar had ice cream cones.
\item[d] {\it They} tasted really good.
\end{enumerate}

\begin{itemize}
\item In a) the expected focus is ``baseball'' (the theme)
\item In b) ``it'' refers to ``baseball'' (CF). ``they'' refers to Alfred and Zohar (AF)
\item The focusing algorithm confirms the CF.
\item In d) ``they'' refers to ``ice cream cones'' in AFL.
\item The focusing algorithm decides that since no anaphor refers to the CF,
  the CF is stacked and ``ice cream cones'' is the new CF (focus movement).
\end{itemize}

We call a basic focusing cycle the cycle that includes : 
\begin{itemize}
\item the focusing algorithm
\item followed by the interpretation of anaphors, 
\item then by the evaluation of the proposed antecedents.
\end{itemize}

\subsection{What needs to be improved in the focusing approach?}

\subsubsection{Intrasentential antecedents}

The focusing approach always prefers the previous sentences' entities as 
antecedents to the current sentences. In fact only previous sentence 
entities are present in the focus registers. Thus phrases of the current 
sentence can not be proposed as antecedents. This problem has already been 
underlined, see \cite{Carter:ACL87} in particular who proposed augmenting the focus
registers with the entities of the current sentence. For example in sentence 
4  while the focus algorithm would propose only ``John'' as an 
antecedent for ``him'', in Carter's method ``Bill'' will also be proposed.
\begin{quote}
4) John walked into the room. He told Bill someone wanted to see {\it him}.
\end{quote}
\subsubsection{Initial Anaphors}
The focusing mechanism fails in the expected focus algorithm when encountering
anaphors occurring in the first sentence of a text, which we call initial anaphors,
such as {\it They} in sentence (1). 
The problem with initial anaphors is that the focus registers cannot be 
initialised or may be wrongly filled if there are anaphors inside the first 
sentence of the text. It is clear that taking the sentence in its classical 
meaning as the unit of processing in the focusing approach, is not suitable 
when sentences are embedded.

We will focus on the mechanisms and algorithmic aspects of the resolution (how 
to fill the registers, how to structure algorithms, etc.) and not on the 
rule aspects, like how IRs decide to choose {\it Bill} and not {\it John} (sentence 4).

\section{Our Solution}
As stated above, embedded sentences include several elementary events (EEs).
EEs are represented as conceptual entities in our work. 
We consider that such successive EEs involve the same context that is 
introduced by several successive short sentences. Moreover, our assumption  
states that when non-PRR anaphors have intrasentential antecedents, they occur in 
embedded sentences. Starting with these considerations, the algorithm is 
governed by the hypotheses expanded below.

\subsection{Main hypotheses}
\begin{description}
\item [First hypothesis]: EE is the unit of processing in the basic focusing cycle.
\end{description}
An EE is the unit of processing in our resolution algorithm instead of the 
sentence. The basic focusing cycle is applied on each EE in turn and not 
sentence by sentence. Notice that a simple sentence coincides with its EE.

\begin{description}
\item [Second hypothesis]: The ``initial''  EE of a well formed first sentence 
does not contain non-PRR pronouns just as an initial simple sentence
cannot. 
\end{description}

For example, in the splitting of sentence 1 into two EEs (see below), EE1 does
 not contain non-PRR pronouns because it is the initial EE of the whole discourse. 

\begin{quote}        
EE1) ``Three of the world's leading advertising groups, Agence Havas S.A. of 
France, Young \& Rubicam of the U.S. and Dentsu Inc. of Japan, said''\\
EE2) ``they are forming a global advertising joint venture.''\\
\end{quote}

\begin{description}
\item [Third hypothesis]: PRR pronouns require special treatment.
\end{description}

PRR could refer to intrasentential antecedents in simple sentences (such as in
those of sentences 3 and 4). An initial EE could then contain an anaphor of
the PRR type. Our approach is to add a special phase that resolves first the
PRRs occurring in the initial EE before applying the expected focusing
algorithm on the same initial EE. In all other cases, PRRs are treated equally
to other pronouns.

This early resolution relies on the fact that the PRR pronouns may refer to the
agent, as in sentence 3, as well as to the complement phrases. However the
ambiguity will not be huge at this first level of the treatment. Syntactic and
semantic features can easily be used to resolve these anaphors.  This relies
also on the fact that the subject of the initial EE cannot be a pronoun (second hypothesis).

Having mentioned this particular case of PRR in initial EE, we now expand on
the whole algorithm of resolution.

\subsection{The Algorithm}
In the following, remember that what we called the basic focusing cycle is the
following successive steps:
\begin{itemize}
\item applying the resolution rules, 
\item applying the focusing algorithm, i.e., updating the focus registers 
\item the evaluation of the proposed antecedents for each anaphor.
\end{itemize}

The algorithm is based on the decomposition of the sentence into EEs and the
application of the basic focusing cycle on each EE in turn and not sentence by
sentence.

The complete steps are given below (see also figure 1):
\begin{description}
\item [Step 1]
        Split the sentence, i.e., its semantic representation, into EEs.\\
\item [Step 2]
        Apply the expected focus algorithm to the first EE.\\
\item [Step 3]
        Perform the basic focusing cycle for every anaphor of all the EEs of the current sentence.\\
\item [Step 4]
        Perform a collective evaluation (i.e., evaluation that involves all 
the anaphors of the sentence), when all the anaphors of the current sentence are 
processed. \\
\item [Step 5]
        Process the next sentence until all the sentences are processed:
        \begin{itemize}
        \item split the sentence into EEs
        \item apply Step 3 then Step 4.
        \end{itemize}
\end{description}

\begin{figure}[htbp]
  \begin{center}
    \leavevmode
    \psfig{file=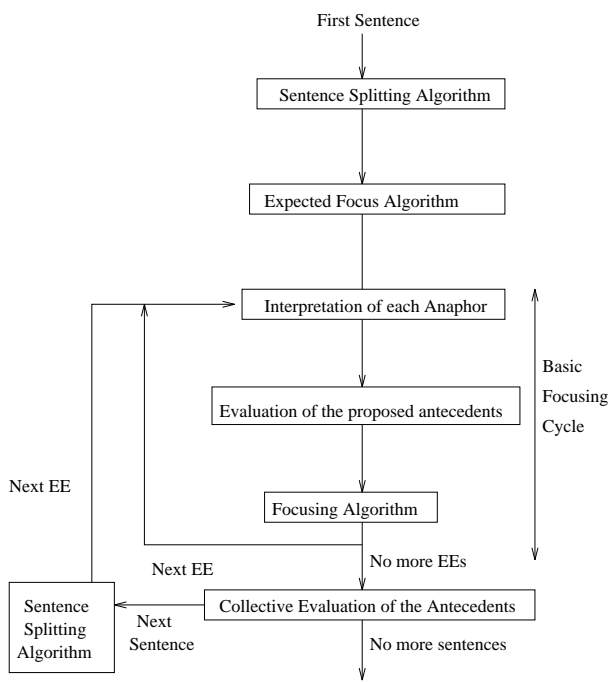,height=9cm}
    \caption{The Algorithm}
  \end{center}
\end{figure}

Main Results :
\begin{enumerate}
\item Intrasentential antecedents are taken into account when applying the
  focusing algorithm. For example, in sentence 1, the intrasentential antecedent {\it Bill} will be taken into account, because EE1 would be processed beforehand by the expected focusing algorithm.
\item The problem of initial anaphors is then resolved. The expected focusing
  algorithm is applied only on the initial EE which must not contain anaphors.
\end{enumerate}

\subsection{Examples and results}
To illustrate the algorithm, let's consider the following sentence :
\begin{quote}
Lafarge Coppee said it would buy 10 percent in National Gypsum, the
number two plasterboard company in the US, a purchase which allows it to be
present on the world's biggest plasterboard market.
\end{quote}
At the conceptual level, there are 3 EEs. They are involved respectively by the 
{\it said}, {\it buy}, and {\it allows} verbs. They correspond respectively to the following
 surface sentences:
\begin{enumerate}
\item[EE1] ``Lafarge Coppee said'' 
\item[EE2] ``it would buy 10 percent in National Gypsum, the number
  two plasterboard company in the US''
\item[EE3] ``a purchase which allows it to be present on the world's biggest
  plasterboard market.''
\end{enumerate}

Consider the algorithm :
\begin{itemize}
\item the expected focusing algorithm is applied to the first EE, EE1, which
 contains non-PRR anaphors.  
\item the other phases of the algorithm, i.e., the basic focusing cycle, are
  applied to the subsequent EEs :
\begin{itemize}
\item EE2 contains only one pronoun {\it it}, which is resolved by the basic
  focusing cycle
\item {\it it} in EE3 will be resolved in the same way.
\end{itemize}
\end{itemize}

The anaphora resolution has been implemented as a part of a conceptual analyser
\cite{Azzam:ACL95a}. It dealt particularly with pronouns. It has been 
tested on a set of 120 news reports. We made two kinds of algorithm evaluations: 
the evaluation of the implemented procedure and an evaluation by hand. For the 
implementation the success rate of resolution was 70\%. The main cases of 
failure are related to the non implemented aspects like the treatment of 
coordination ambiguities and the appositions, or other anaphoric phenomena, like ellipsis.

For the second evaluation which concerns the real evaluation of the approach,i.e.,
 without going into the practical issues concerning  implementation, the 
success rate was  95\%. The main cases of failure were due to the cases that
were not considered by the algorithm, like for example the pronouns 
occurring before their antecedents , i.e., cataphors. Such cases occur for example in sentences 5 and 6 pointed out by Hobbs \cite{Hobbs:ACL85} to discuss the cases that are not handled easily in the literature.

\begin{quote}
5) Mary sacked out in {\it his} apartment before {\it Sam} could kick her out.\\
6) Girls who {\it he} has dated say that {\it Sam} is charming.
\end{quote}

Our algorithm fails in resolving {\it his} in 5, because the algorithm searches only for the entities that preceed the anaphor in the text. The same applies for {\it he} in 6. However improving our algorithm to process classical cases of cataphors, such as those in sentence 6, should not require major modifications, only change in the order in which the EEs are searched through.

For example, to process pronouns of the sentence 6 split into two EES (see below), the algorithm must consider EE2 before EE1. This means applying the step 2 of the algorithm to EE2, then step 3 to EE1. The sentence 5 should require specific treatment, though.

\begin{quote}
EE1) ``that {\it Sam} is charming''\\
EE2) ``Girls who {\it he} has dated say''
\end{quote}

Hobbs also pointed out the cases of ``picture noun'' examples, as in sentences 7 and 8:
\begin{quote}
7) John saw a picture of him. \\
8) John's father's portrait of him.
\end{quote}
In 7 our algorithm is successful, i.e., it will not identify {\it him} 
with {\it John} because of our previous assumption (section 2.2). However our algorithm would fail in 8 because the non-PRR pronoun {\it him} 
could refer to {\it John} which occurs in the same EE. 

Notice that Hobbs' \cite{Hobbs:ACL85} remark that ``the more deeply the pronoun is embedded and 
the more elaborate the construction it occurs in, the more acceptable the non reflexive'' is consistent with our assumption.

For example in the embedded sentence 9 where either the reflexive (himself) or non
reflexive pronouns (him) may be used, it is more natural to make use of {\it him}.
\begin{quote}
9) John claimed that the picture of him hanging in the post office was
a fraud.
\end{quote}
\section{The Conceptual Level}
We comment here on the main aspects of the conceptual analysis that are related 
to the anaphora resolution process. They concern mainly the way of splitting 
embedded sentences and the problems of determining the theme and of managing the 
other ambiguities and the several readings.

The conceptual analyser's strategy consists of a continuous 
step-by-step translation of the original natural language sentences into 
conceptual structures (CS hereafter). 
This translation uses the results of the syntactic analysis 
(syntactic tree). It is a progressive substitution of the NL terms located in 
the syntactic tree with concepts and templates of the conceptual 
representation language. Triggering rules are evoked by words of the sentence 
and allow the activation of well-formed CS templates when the 
syntactico-semantic filter is unified with the syntactic tree. The values 
caught by the filter variables are the arguments of the CS roles, i.e., 
they fill the CS roles. If they are anaphors, they are considered to be 
unbound variables and result in unfilled roles in the CS. The anaphora 
resolution aims therefore at filling the unfilled roles with the 
corresponding antecedents.

\subsection{Splitting into EE}
The splitting of a sentence in EE is done on the corresponding CS. A minimal 
CS is a template comprising a predicate that identifies the basic type of 
the represented event and a set of roles or predicate cases. \\
For example, 
the sentence ``to say that they agree to form a joint venture'' is represented, 
in a simplified way, with three templates, corresponding to the predicates: 
\begin{itemize}
\item {\it move} information (from ``to say''),
\item {\it produce} an agreement (from ``to agree''),
\item {\it produce} a joint venture (from ``to form'').
\end{itemize}
Given that one template at the semantic level represents an elementary event, 
the splitting is implicitly already done when these templates are created in 
the triggering phase. Indeed, the syntactico-semantic filter of the triggering 
rules takes into account the semantic features of words (mainly verbs) for 
recognising in the surface sentence those that are able to trigger an elementary 
event.

\subsection{Determining the theme}
Gruber and Anderson characterise the theme as follows: if a verb describes a 
change to some entity, whether of position, activity, class or possession, then 
the theme is the changed entity, \cite{Gruber:ACL76} and \cite{Anderson:ACL77}. As 
Carter \cite{Carter:ACL87} demonstrated, this definition of Gruber and Anderson 
is sufficient to apply the focusing mechanism. This assumption is 
particularly apt when we dispose of a conceptual representation. Indeed, 
to determine the thematic roles, we established a set of thematic rules that 
affect for a given predicative occurrence, its thematic functions according to 
the predicate type, the role type and the argument's semantic class.

\subsection{Managing other ambiguities}
An important aspect appears when one designs a concrete system, namely how to 
make other disambiguation processes cohabit. In the conceptual analyser, the general 
disambiguation module (GDM) deals with other ambiguities, like  
prepositional phrase attachment. It coordinates the treatment of different 
kinds of ambiguities. 
This is necessary because the conceptual structures (CS) on which the rules 
are performed could be incomplete because of other types of ambiguities not 
being resolved.\\
For example, if the CF of the sentence is a PP object that is 
not attached yet in the CS the thematic rules fail to fill the CF. The GDM 
ensures that every disambiguation module intervenes only if previous 
ambiguities have already been resolved. The process of co-ordinating ambiguity 
processing is fully expanded in \cite{Azzam:ACL95b}.

\subsection{Multiple readings}
When dealing with ambiguities, another important aspect is managing multiple 
readings. At a certain point when the GDM calls the anaphora module to deal 
with a given anaphor, the status of the conceptual analysis could be 
characterised by the following parameters :
\begin{itemize}
\item The set of conceptual structures for the current reading Ri on which the
  resolution is performed, given that several readings could arise from
  previous ambiguity processing.
\item The set of conceptual structures of the current sentence Si where the
  anaphor occurs;
\item The set of conceptual structures of the current elementary event EEi
  where the anaphor occurs after the Si splitting.
\item The state of the focus (content of the registers), SFi
\end{itemize}
The main assumption is that the anaphora resolution algorithm always applies 
to a single state, (Ri, Si , EEi, SFi) when resolving a given anaphor 
(Step 3) :
\begin{enumerate}
\item[a] If several antecedents are still possible after the individual evaluation
  of the anaphor, Ri is then duplicated, in Rij, as many times as there are
  possibilities.
\item[b] When performing the collective evaluation of all Si anaphors, every
  inconsistent Rij is suppressed.
\item[c] The result is a set of readings (Rij, Sj , EEj, SFi).
\end{enumerate}

\section{Conclusion}
We have proposed a methodology to resolve anaphors occurring in embedded 
sentences. The main idea of the methodology is the use of other kinds of 
restrictions between the anaphor and its antecedents than the syntactic ones. 
We demonstrated that anaphors with intrasentential antecedents are closely related 
to embedded sentences and we showed how to exploit this data to design the anaphora resolution methodology. Mainly, we exploited Sidner's focusing mechanism, refining the classical unit of processing, that is the sentence, to that of the elementary event. The algorithm has been implemented (in Common Lisp, Sun Sparc) to deal with pronouns as a part of a deep analyser. The main advantages of the proposed algorithm is that it is independent from the knowledge representation language used and the deep understanding approach in which it is integrated. 
Thus, it could be set up in any conceptual analyser, as long as a semantic 
representation of the text is available. Moreover Sidner's approach does not 
impose its own formalisms (syntactic or semantic) for its application. 
 The improvment of the proposed algorithm requires dealing with special cases of anaphors such as cataphors and also with specific cases which are not easily handled in the literature. For example, we saw that a solution to processing cataphors could be to reconsider the order in which the conceptual structures (elementary events beforehand) are searched. 

\section{Acknowledgements}
This work has been supported by the European Community Grant LE1-2238 (AVENTINUS project).

\bibliographystyle{fullname}

\end{document}